\documentclass[10pt,twocolumn,letterpaper]{article}

\usepackage{cvpr}
\usepackage{times}
\usepackage{epsfig}
\usepackage{graphicx}
\usepackage{amsmath}
\usepackage{amssymb}
\usepackage{graphicx}
\usepackage{subcaption}
\usepackage{float}
\usepackage{appendix}
\usepackage{longtable}

\usepackage[pagebackref=true,breaklinks=true,colorlinks,bookmarks=false]{hyperref}

\cvprfinalcopy % *** Uncomment this line for the final submission

\def\cvprPaperID{****} % *** Enter the CVPR Paper ID here

\ifcvprfinal\fi
\begin{document}

%%%%%%%%% TITLE
\title{Efficient LLM Context Distillation}

\author{
Rajesh Upadhayaya\textsuperscript{1}, Manish Raj Osti\textsuperscript{1}, Zachary Smith\textsuperscript{1}, Christopher Kottmyer\textsuperscript{1} \\
\textsuperscript{1}School of Computer Science, Georgia Institute of Technology, Atlanta, USA \\
{\tt\small upadraj@gatech.edu, mrosti@gatech.edu, zsmith73@gatech.edu, ckottmyer3@gatech.edu}
}

\maketitle
%\thispagestyle{empty}

%%%%%%%%% ABSTRACT
\begin{abstract}
   Large Language Models (LLMs) demonstrate proficiency across diverse tasks but often require targeted adaptations for specific applications. Various methods have been proposed to facilitate this adaptation, including few-shot fine-tuning, in-context learning, and context distillation. This paper specifically investigates context distillation — a method that extends the utility of task-specific examples by internalizing them, thus augmenting the example set accessible for model inference. We conduct a comparative analysis of context distillation with in-context learning (ICL) and few-shot fine-tuning (FT), aiming to ascertain the efficacy of context distillation in adapting models using minimal in-context examples. Employing matched datasets from Mobach, our experiments leverage OPT models of various sizes. The results indicate that context distillation effectively adapts models, with student models attaining comparable in-domain and out-of-domain accuracies to in-context learning. Although context distillation surpasses ICL in out-of-domain generalization, it does not achieve the performance levels of FT. However, the reduced dataset size and computational demands position context distillation as a viable alternative, especially for smaller datasets. Overall, this study presents context distillation as an efficient and potent method for customizing LLMs to specific tasks.
\end{abstract}

%%%%%%%%% BODY TEXT
\section{Introduction}

Large language models (LLM) excel at knowledge extraction and reasoning, but often require adaptation to individual tasks. There are several proposed methods to perform this adaptation including, but not limited to, few-shot fine-tuning, in-context learning, and context distillation. Each method has its own advantages and disadvantages. For example, few-shot fine-tuning (FT) requires substantial amounts of task specific data, which poses a challenge when labeled examples are scarce. In-context learning (ICL) attempts to alleviate FT data needs by providing fewer examples through the query prompt used during inference. However, LLM have a constrained context window that is both consumed by task examples and limit the number of examples that models can learn from. Given this constrained context window, context distillation (CD) extends accessible task-specific examples by internalizing them, greatly increasing the number of available examples outside of the query prompt \cite{anthropic2021}. This not only limits the number of task examples these models can learn from simultaneously but also affects their ability to integrate and recall relevant information across different parts of the text. Context distillation (CD) addresses these limitations by allowing a model to internalize and condense key information from task-specific examples. This process effectively extends the usable information beyond what is immediately present in the query prompt, increasing the number of examples the model can learn from and utilize, without being directly constrained by the size of the context window.

In this paper, we explore context distillation by directly comparing our results to the ICL results in Mosbach et al. \cite{mosbach2023fewshot}. Mosbach et al. show that ICL is a viable task adaptation method despite it under-performance relative to FT. A plausible explanation for this under-performance is that FT learns from many training examples whereas ICL learns from only a few in-context examples provided during inference. This results in ICL having less data for task generalization. By contrast, context distillation strikes a balance between the large number of examples required for FT and the small number of examples that fit into the context window of ICL. Additionally, Snell et al. shows that CD outperforms direct gradient descent learning \cite{snell2022learning}. Thus, it is hypothesized that context distillation should perform better than ICL without requiring as many training examples as FT.

Our goal is to demonstrate that context distillation can be performant by using only a few in-context examples. This not only signifies efficient training data utilization but also showcases task-specific improvements relative to conventional fine-tuning methods trained on small datasets. When combined with the efficiency gains of low-rank adaptation layers (LoRa), context distillation becomes an efficient training method for learning task specific adaptations and provides increased flexibility. To accomplish this a fixed reference model with task-specific LoRa layers is used during inference time.

In terms of data, we employ matched datasets similar to those utilized in previous studies, ensuring consistency and comparability. Our datasets encompass a range of tasks, including natural language inference (NLI) and paraphrase identification. Datasets are sourced from widely available repositories such as \href{https://huggingface.co/}{Hugging Face}. By leveraging these datasets, we provide a comprehensive evaluation of context distillation on a diverse set of tasks and domains.

We examine how context distillation mitigates limitations in traditional fine-tuning by comparing our results to previous studies. We provide insights into the comparative efficacy of context distillation relative to FT on task adaption.

%-------------------------------------------------------------------------
%-------------------------------------------------------------------------

\section{Approach}

Overall, our approach was to implement context distillation by adapting methods used in Mosbach et al. \cite{mosbach2023fewshot}. This includes referencing and modifying their code to adapt it to our teacher-student training procedure and to retrieve their datasets for a comparative analysis. Adaption in this context means selectively using their code, changing it to be more efficient and flexible, and even fixing bugs encountered in their code. We note that the original code base was bloated and badly abstracted. By the end of the project, we completely rewrote sections of their code and used the original code to selectively validate our code. A significant part of validation was confirming we reconstructed their datasets to guarantee our results were comparable to theirs. For example, we needed to confirm that they binarized one of the datasets and make sure we replicated it in our dataset. Python was used to access PyTorch, a deep learning framework, as well as Hugging Face Transformers \cite{wolf-etal-2020-transformers} and Datasets \cite{lhoest-etal-2021-datasets}.

%-------------------------------------------------------------------------

\subsection{Datasets}

The datasets were consistent with those used by Mosbach et al. \cite{mosbach2023fewshot}. Two common natural language processing tasks were used: natural language inference (NLI) and paraphrase identification. For the NLI task, MNLI \cite{mnli2018} and RTE \cite{rte2006} where used for our in-domain datasets and the lexical overlap subset of HANS \cite{mccoy-etal-2019-right} was used as our out-of-domain (OOD) dataset. MNLI is binarized by removing neutral examples matching the labels in RTE and HANS datasets. For the paraphrase identification task, QQP \cite{sharma2019natural} was used as our in-domain dataset and PAW-QQP \cite{zhang-etal-2019-paws} as our OOD. All datasets were configured to use yes labels for entailment or paraphrase and no labels otherwise.

These datasets are used for natural language processing (NLP) and serve as common benchmarks. They are widely available through platforms such as \href{https://huggingface.co/}{Hugging Face}. Each in-domain dataset comes with training and validation subsets. The training sets where first sampled to create in-context examples. They are then randomly sampled for our query examples which are used for inference. Care was taken to ensure that the in-context and query examples did not overlap.

Regarding data preparation, both the context distillation teacher and student models are structurally the same, decoder only transformers, and so the expected inputs are tokenized strings. The output, after applying softmax, is a series of tokens that after using a model-specific tokenizer for decoding results in a string representation. Additionally, no pre- or post-processing was required since the datasets are used as NLP benchmarks.

%-------------------------------------------------------------------------

\subsection{Models}

All experiments were run using 4 OPT models \cite{zhang2022opt} - 125 million, 350 million, 1.3 billion, and 2.7 billion parameters. Using the OPT family of models guarantees that the models where trained on the same dataset. This isolates the impact of model size from a possible confounding factor of the training datasets used.

%-------------------------------------------------------------------------

\subsection{Context Distillation Setup}

The context distillation was set-up using both a teacher and a student model tasked to infer a label for an inference request. The teacher model receives a string of in-context examples followed by an inference request without a label. The student model receives only the inference request. Both models generate an answer and the difference between the two answers, via KL-divergence loss, is used to update the student model weights. The teacher model weights stay frozen.

To generate training contexts for the teacher model, we randomly sampled the in-domain training set, varying the number of context examples $n \in \{2, 16, 32\}$ . This process is repeated four times for each $n$, resulting in four unique context example sets per $n$. However, for the RTE task, $n = 32$ consistently exceeded the context window, so we could not run it. All in-context examples and inference requests were formatted according to the pattern in Table \ref{tab:patterns}.

During training, each set of context examples served as the training context for a single run per model. During that run, we fine-tuned the student model on 32 inference requests, which were randomly sampled from the in-domain training set. To prevent data leakage, none of the 32 inference requests overlapped with the teacher's context examples.

\begin{table*}
\begin{center}
\begin{tabular}{|l|p{1.5cm}|p{7cm}|l|l|}
\hline
Type & Dataset(s) & Pattern text & Answer prefix & Target tokens \\
\hline
General pattern & All & Premise: \{premise\} \textbackslash n Hypothesis: \{hypothesis\} \textbackslash n  &  Label: \{label\} & Yes, No\\
\hline
Teacher pattern & MNLI, RTE & Think logically. Are the following sentences examples of entailment, yes or no?\textbackslash n \{context example $n_1$\}\textbackslash n\textbackslash n \{context example $n_2$\}\textbackslash n\textbackslash n ...\{context example $n_n$\}\textbackslash n\textbackslash n \{inference example\} & Label: & Yes, No \\
\hline
Teacher pattern & QQP & Think logically. Are the following sentences duplicates or paraphrases of each other, yes or no?\textbackslash n \{context example $n_1$\}\textbackslash n\textbackslash n \{context example $n_2$\}\textbackslash n\textbackslash n ...\{context example $n_n$\}\textbackslash n\textbackslash n \{inference example\} & Label: & Yes, No \\
\hline
Student pattern & MNLI, RTE & Are the following sentences examples of entailment, yes or no?\textbackslash n \{inference example\} & Label: & Yes, No \\
\hline
Student pattern & QQP & Are the following sentences duplicates or paraphrases of each other, yes or no?\textbackslash n \{inference example\} & Label: & Yes, No \\
\hline
\end{tabular}
\end{center}
\caption{Patterns used for context distillation. All context examples and inference examples are formatted using the general pattern. Context examples include the answer after the answer prefix. The inference examples do not.}
\label{tab:patterns}
\end{table*}

%-------------------------------------------------------------------------

\subsection{Fine-tuning}

The same fine-tuning process was used for all models and tasks. Each context distillation run used an in-domain dataset for training and was validated using both the in-domain and OOD datasets for that task. The last token of the teacher and student model outputs were assumed to represent the label. KL divergence between the teacher and student output was used to update the student LoRa layers' weights.

Since both the teacher and student model are the same size model, LoRa adaptors where utilized. The implementation of the low-rank adaptors require that the pre-trained model parameters are frozen, creating a reference model. A low-rank adaptation matrix layer is then applied on top of the reference model. During fine-tuning, the adaptation matrix, as represented by a pair of rank decomposition matrices, is updated. This drastically reduces the number of updated parameters for each training step. The frozen reference model was used as our teacher model.

To make our results comparable to Mosbach et al. we used their hyperparameters \cite{mosbach2023fewshot}. These parameters were specifically recommended by Mosbach et al. \cite{mosbach2021on} and are referenced in this Table \ref{tab:hyperparams}

\begin{table*}
\begin{center}
\begin{tabular}{|l|l|}
\hline
Hyperparameter & Value \\
\hline
Optimizer & AdamW \\
Learning rate & $10^{-5}$ \\
Learning rate schedule & linear warmup then constant \\
Warmup ratio & 10\% of total steps \\
Weight decay & 0.0 \\
Dropout & 0.1 \\
Batch size & 32 \\
Epochs & 20 \\
Total steps & $\frac{\#samples}{batch size} * epochs$ \\[2mm]
\hline
\end{tabular}
\end{center}
\caption{Hyperparameters used in training}
\label{tab:hyperparams}
\end{table*}

%-------------------------------------------------------------------------

\subsection{Evaluation}

For validation, we randomly sampled 100 examples from both the in-domain validation and OOD validation datasets. A model was then evaluated on the both the in-domain validation set followed by the OOD validation dataset that matched the in-domain dataset's task.

After completing all of the Context Distillation experiments - each model size, trained on each in-domain dataset, is validated using the student model on both corresponding in-domain and OOD datasets. The in-domain validation and OOD results are compared to Mosbach et al. \cite{mosbach2023fewshot} to assess the efficacy of our approach. Snell et al. \cite{snell2022learning} shows that Context Distillation out performs direct gradient descent learning on the T5 model \cite{raffel2023exploring}. Therefore, we felt that our approach should reasonably improve on the ICL results.

A direct comparison of in-context learning vs context distillation across our 4 model sizes has not been performed previously. We focused on filling this knowledge gap by conducting experiments that compare CD to ICL results.

%-------------------------------------------------------------------------
%-------------------------------------------------------------------------

\section{Experiments}

As outlined in the approach section, we conducted a series of teacher-student context distillation experiments. Paraphrasing what was presented, we evaluate CD by training on three benchmark datasets - MNLI and RTE for the NLI task and QQP for the paraphrase task. For each dataset we construct teacher contexts of three sizes, $n \in \{2, 16, 32\}$. For each context size we sample 4 unique in-context example sets. Each of these example sets equates to a CD training run. We then trained OPT models of 4 different sizes on each training run by sampling 32 training examples and stripping to the label to use as an inference point. 

We measure success by comparing the performance of student models (trained with CD) to teacher models (untrained) and previously established methods (ICL and FT) on held-out validation datasets both in-domain and out-of-domain.

%-------------------------------------------------------------------------
%-------------------------------------------------------------------------

\section{Results}

%-------------------------------------------------------------------------

\subsection{Context Distillation}

Our context distillation experiments reveal several notable findings. The average accuracy of all four runs for each dataset and model are presented in Table \ref{tab:accuracy}. Figures presenting the results of all runs for all teacher context lengths can be found in the appendix.

In terms of success, context distillation proved useful. From Table \ref{tab:accuracy}, one can see that for all scenarios the student model achieved comparable in-domain and OOD accuracy to the teacher's in-domain accuracy. For the MNLI and RTE datasets the student model only had a slight reduction in OOD accuracy compared to its in-domain accuracy. However, QQP presents a peculiar behavior by performing better on OOD data than in-domain. The small parameter models, 125m and 350m, performed better after CD on in-domain validation. This shows the benefit of the student model seeing more examples through context distillation than the teacher. The teacher having seen only the 16 in context examples, while the student benefits from those and the 32 inference examples used to update the LoRa weights during tuning. For the larger models, 1.3b and 2.7b, the teacher performed better than the student, likely a result of the knowledge ingrained with more parameters.

Interestingly, context distillation seems to alleviate the impact of model size on the capability of the model. Where the teacher accuracy increased with model size the student accuracy was more stable across the sizes. The indicates CD may be a valuable in enabling smaller models to perform as well as larger models on specific task.

Finally, CD did not overfit on the training data. As evidenced by the good performance of the context distilled student model on OOD validation set. Using LoRa layers with a frozen reference model helped to prevent overfitting and catastrophic forgetting as well.

%-------------------------------------------------------------------------

\subsection{Comparison of Task Tuning Methods}

\begin{figure*}
    \centering
    \begin{subfigure}[t]{0.8\textwidth}
        \includegraphics[width=\textwidth]{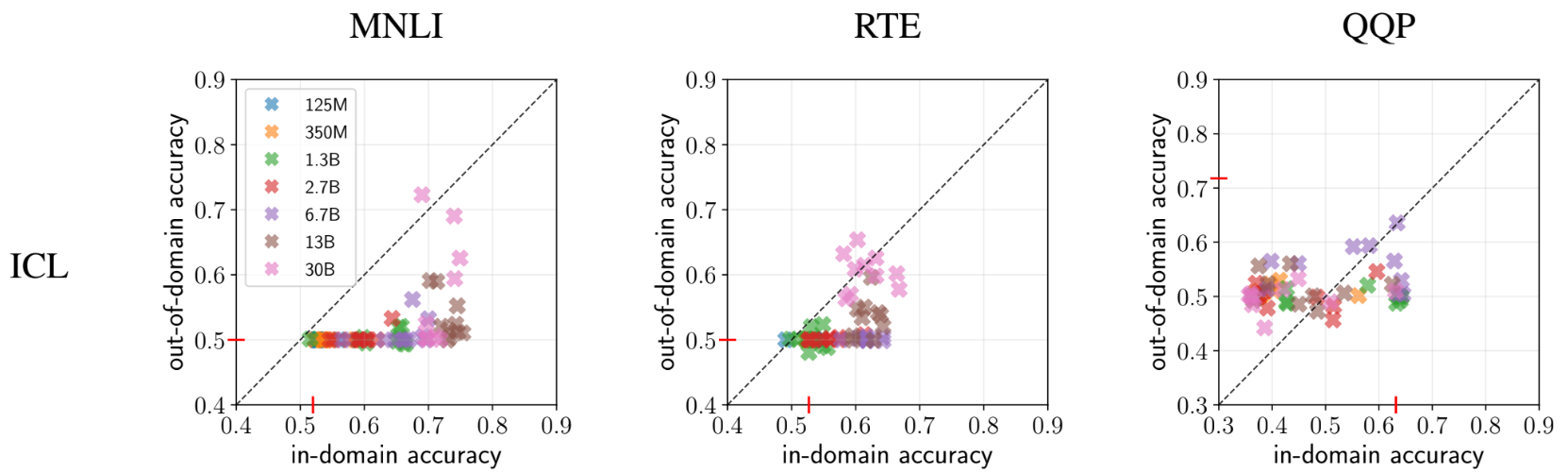}
        \caption{}
        \label{fig:icl-cd-a}
    \end{subfigure}
    \begin{subfigure}[t]{0.8\textwidth}
        \includegraphics[width=\textwidth]{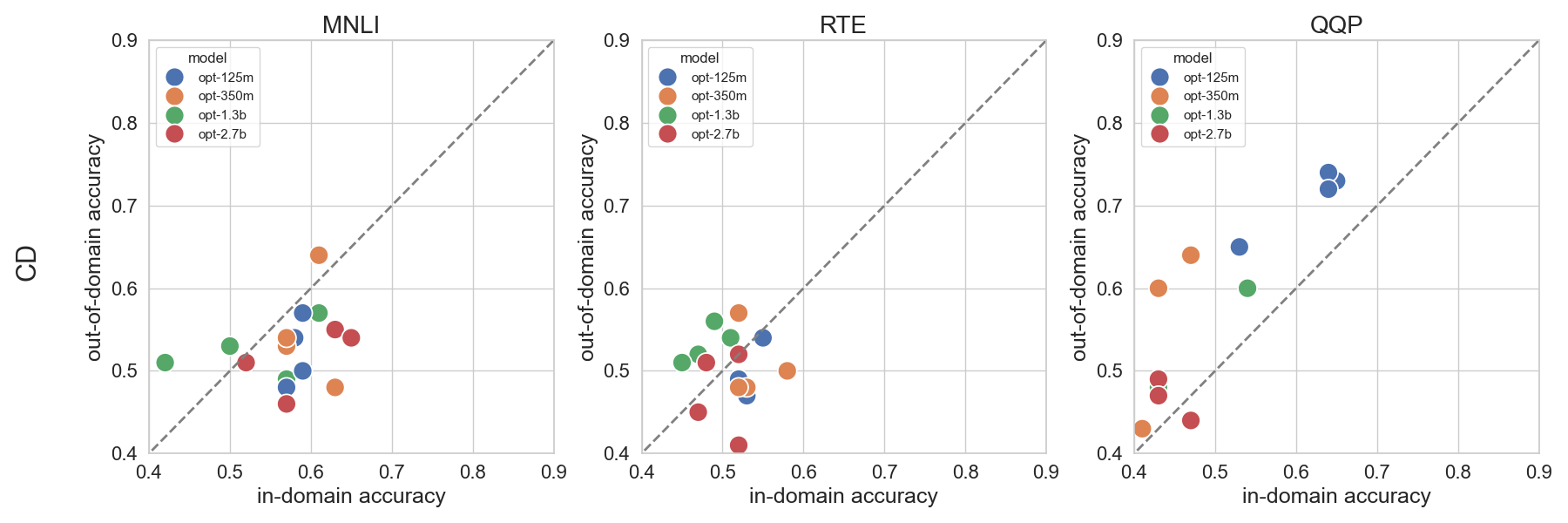}
        \caption{}
        \label{fig:icl-cd-b}
    \end{subfigure}
    \caption{Sub-figure (a) comprises the results of ICL as published by Mosbach et al. \cite{mosbach2023fewshot}. Sub-figure (b) are the results of CD from our experiments. Both figures represent the scenario using $n=16$ context examples.}
    \label{fig:icl-cd}
\end{figure*}

In comparing the different task tuning methods, our findings suggest that CD offers notable quality advantages over ICL. The results in Figure \ref{fig:icl-cd} show our experimental outcomes in comparison to Mosbach et al. \cite{mosbach2023fewshot} for the $n=16$ context examples training runs. CD shows comparable in-domain performance, but clearly improves on the OOD performance across datasets and model sizes. 

On the other hand, CD does not perform as well as FT. However, with CD the training dataset size and computation required to compute are drastically reduced. Of note, the improvement achieved through FT is less drastic on the million parameter models that on the billion plus parameter models.

%-------------------------------------------------------------------------
% TABLES AND FIGURES
% Accuracy
\begin{table*}
\begin{center}
\begin{tabular}{|l|l|l|l|l|}
\hline
Dataset & Model size & Accuracy teacher & Accuracy in-domain & Accuracy OOD \\
\hline
MNLI & 125m & 0.547 & \textbf{0.583} & 0.523 \\
     & 350m & 0.531 & \textbf{0.595} & 0.548 \\
     & 1.3b & \textbf{0.609} & 0.525 & 0.525 \\
     & 2.7b & \textbf{0.664} & 0.593 & 0.515 \\ 
\hline
RTE  & 125m & 0.438 & \textbf{0.530} & 0.518 \\
     & 350m & 0.508 & \textbf{0.538} & 0.508 \\
     & 1.3b & 0.516 & 0.480 & \textbf{0.533} \\
     & 2.7b & \textbf{0.547} & 0.498 & 0.473 \\ 
\hline
QQP  & 125m & 0.625 & 0.615 & \textbf{0.710} \\
     & 350m & 0.406 & 0.420 & \textbf{0.490} \\
     & 1.3b & 0.406 & \textbf{0.430} & 0.415 \\
     & 2.7b & \textbf{0.438} & 0.428 & 0.433 \\
\hline
\end{tabular}
\end{center}
\caption{Comparison of accuracy for the teacher model on in-domain data and post-tuning student model on in-domain and OOD data. Results represent the scenario using $n=16$ context examples. Best result per model size in bold.}
\label{tab:accuracy}
\end{table*}

% ICL vs CD

%-------------------------------------------------------------------------
%-------------------------------------------------------------------------

\section{Experience}

During our research, we anticipated writing code and uncovering bugs we generated. We did not anticipate how much code we would write and refactor. The infrastructure was not trivial to set-up and involved running 8+ instances on Google Colab in parallel with T4, L4 and A100 GPUs (2+ students simultaneously). We spent a significant amount of time reverse-engineering the code from Mosbach et al \cite{mosbach2023fewshot}. We wanted to be through and have comparable results relative to our benchmark. This meant replicating any transformation to their dataset. For example, the benchmark binarized the MNLI dataset. Understanding their experiment set-up felt like trial and error, we spent time in several group calls discussing the finer details of how they set-up context examples and inference requests. This resulted in us re-writing our entire code base at least twice. In each rewrite, we had to fix bugs including ones that we found in the benchmark code base itself. When writing training code our team spent considerable time understanding the tokenizer and trying to retrieve labels. We ended up confirming that the benchmark didn't use labels provided within the dataset and had to recast them to: yes and no labels. 

During training, we reduced the benchmark hyper-parameters, because we had significantly less GPU compute and memory available to us. Our reduced hyper-parameters diminished GPU compute needed to 5\% of the benchmark. This reduced training time from multiple days to 4-5 daily sessions over the weekend. While running models, one teammate modified our DL model with LoRa adapter significantly cutting both compute and memory requirements. Even after these changes, the OPT 2.7b model struggled to complete a run on the A100 often running out of memory. For the 32 context RTE model, the context was too large and so it didn't run. We had to cut it from our experiments. After generating the results for all 128 runs, we aggregated the results from JSON into CSV file. One of our teammates then thoroughly analyzed the results to generate the figures and charts presented in this paper. Overall, the team put in a valiant effort and overcame many hurdles to create this paper and we are happy with the result.

%-------------------------------------------------------------------------
%-------------------------------------------------------------------------

\section{Conclusion}

Models trained with context distillation internalize context examples allowing them to be used during inference. This greatly increases examples that a model learns from avoiding the context window limit. In our paper, context distillation performed well on the out-of-domain dataset and had comparable results to in-context learning on the in-domain dataset. Our study did not tease out the influence of model size on performance. We believe this is due to our limited sample size: 4 runs per model per dataset per context length. We plan on rectifying this with future runs, but note that our initial 120 runs required significant compute. This compute includes upwards of 2 A100s and a minimum of 4 GPUs run in parallel over multiple days. The computational requirement was noticeable for the OPT 2.7b model, which caused the 40 GB memory A100 GPU on multiple occasions to run out of memory for both the MNLI and RTE datasets. Overall, the results were promising with CD showing advanced performance over ICL especially for out-of-domain generalization.

Context distillation is a state-of-the-art training regime and is used in multiple domains. We thought about using context distillation for LLM code generation and internalizing code repositories. We are considering a future ablation study where we investigate performance of an amalgamated model: LoRa, knowledge distillation, and context distillation and then selectively remove components to measure their impact on performance. As for our current research presented in this paper, we would like to run the models to generate more samples so that we can properly tease out the effect of model size on performance. An alternative approach is using 128 inference requests instead of 32 during training. During hyper-parameter tuning, this change looked promising as several large models had over 70\% accuracy for a few tasks. We chose 32 inference requests due to computational constraints. Snell et al. \cite{snell2022learning} performs context distillation with a scratchpad and we would love to replicate that work. We also wanted to conduct an experiment by providing the student model with a small in-context dataset to measure its performance and we also thought about using context distillation on instruction-fine-tuned models.

%-------------------------------------------------------------------------
%-------------------------------------------------------------------------

%-------------------------------------------------------------------------
%-------------------------------------------------------------------------

\twocolumn
\clearpage
{\small
\bibliographystyle{ieee_fullname}
\bibliography{relu-ranger}
}

\onecolumn
\clearpage
\appendix\label{app:cd-exp}
\section{Context distillation experiments}
\begin{figure*}[h!]
    \centering
    \begin{subfigure}[c]{0.8\textwidth}
        \includegraphics[width=\textwidth, height=5cm]{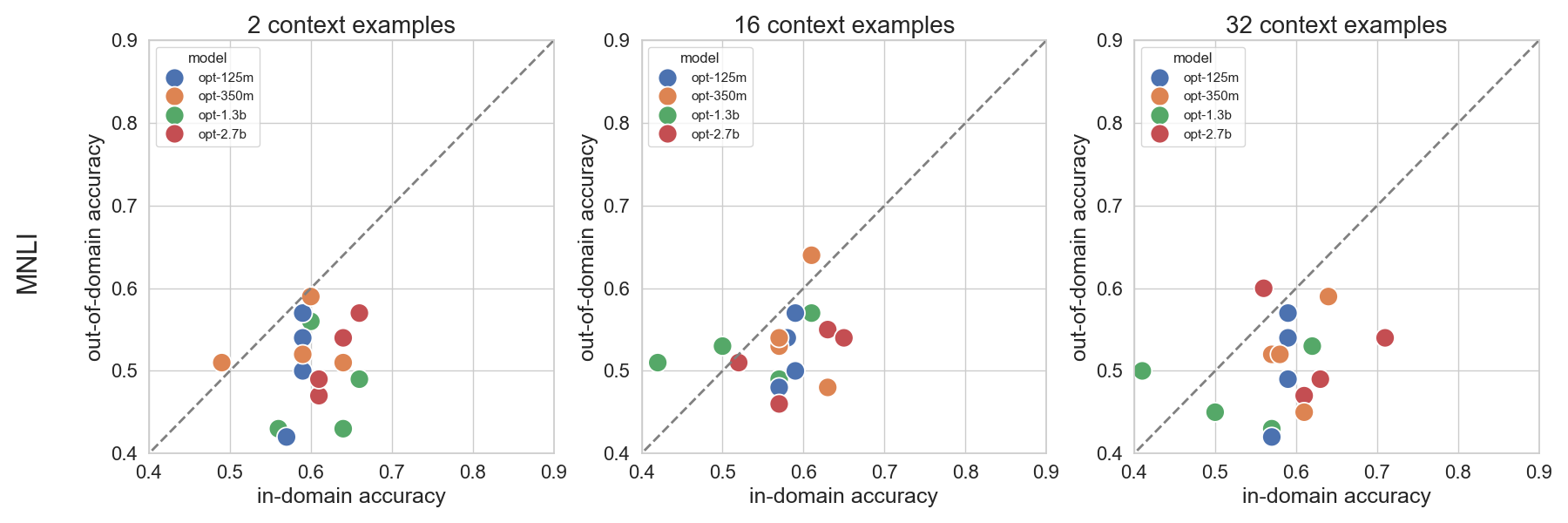}
        \caption{}
        \label{fig:mnli-cd}
    \end{subfigure}
    \begin{subfigure}[c]{0.5\textwidth}
        \includegraphics[width=\textwidth, height=5cm]{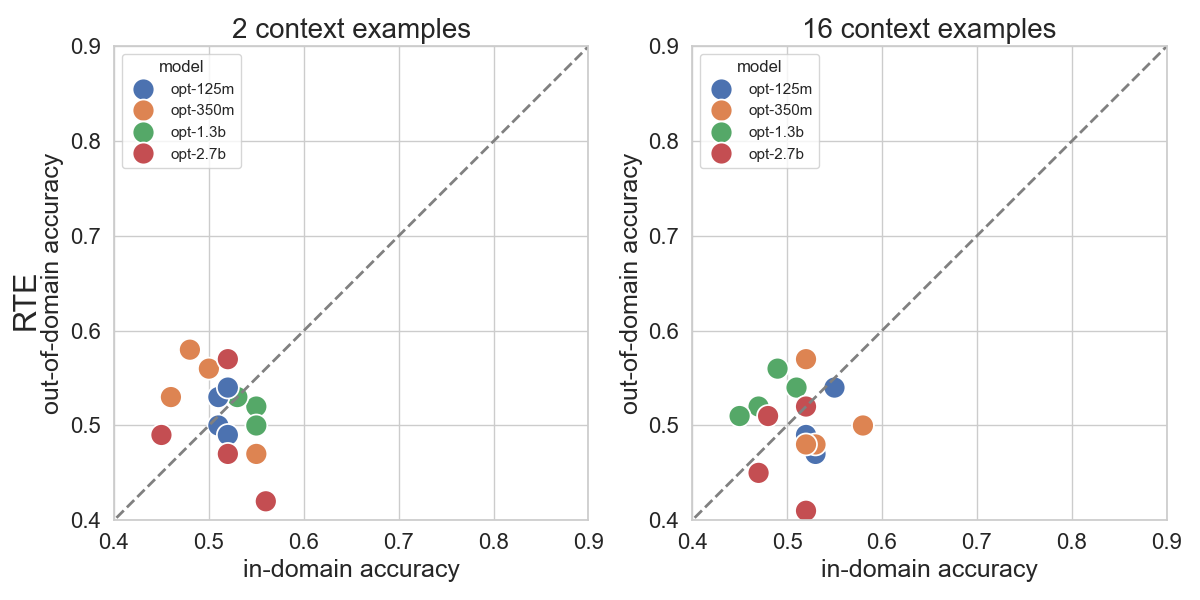}
        \caption{}
        \label{fig:rte-cd}
    \end{subfigure}
    \begin{subfigure}[c]{0.8\textwidth}
        \includegraphics[width=\textwidth, height=5cm]{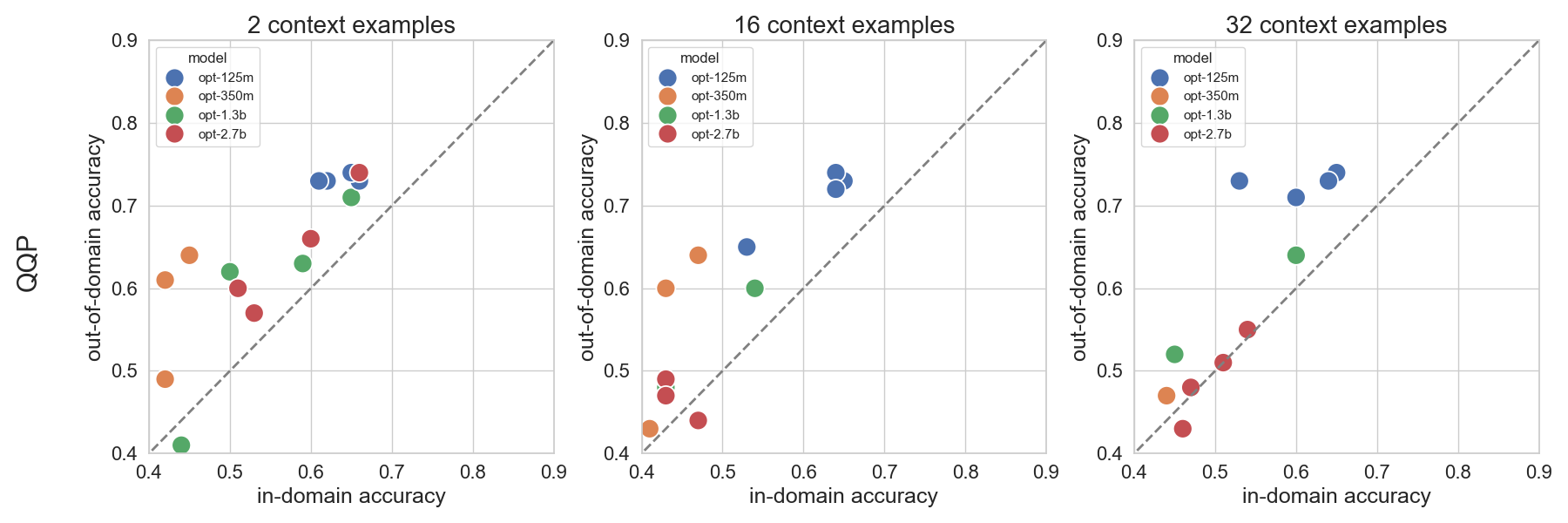}
        \caption{}
        \label{fig:qqp-cd}
    \end{subfigure}
    \caption{Exploring the effect of CD on model quality. Shown are the accuracy of the context distilled student model.}
    \label{fig:cd-exp}
\end{figure*}
\end{document}